\pdfoutput=1

\documentclass[11pt]{article}

\usepackage[preprint]{acl}

\usepackage{times}
\usepackage{latexsym}

\usepackage[T1]{fontenc}

\usepackage[utf8]{inputenc}

\usepackage{microtype}

\usepackage{inconsolata}
\usepackage{multirow}
\usepackage{xcolor}
\usepackage{color}
\usepackage{colortbl}
\usepackage{booktabs}
\usepackage{graphicx}
\usepackage[most]{tcolorbox}
\usepackage{hyperref}
\usepackage{fontawesome}

\newcommand{\prompt}{SKP}
\newcommand{\benchmark}{SUBARU}

\definecolor{green0}{RGB}{190, 245, 163}
\newcommand{\SR}{\cellcolor{green0}}
\definecolor{green1}{RGB}{118, 189, 83}
\newcommand{\UR}{\cellcolor{green1}}
\definecolor{green2}{RGB}{63, 137, 26}
\newcommand{\SSR}{\cellcolor{green2}}

\title{Have We Designed Generalizable Structural Knowledge Promptings?\\ Systematic Evaluation and Rethinking}

\author{
    Yichi Zhang$^\spadesuit$$^\diamondsuit$,
    Zhuo Chen$^\spadesuit$$^\diamondsuit$,
    Lingbing Guo$^\spadesuit$$^\diamondsuit$, 
    Yajing Xu$^\spadesuit$$^\diamondsuit$,
    Shaokai Chen$^\spadesuit$$^\diamondsuit$,\\ 
    \textbf{Mengshun Sun}$^\clubsuit$,
    \textbf{Binbin Hu}$^\clubsuit$,
    \textbf{Zhiqiang Zhang}$^\clubsuit$,
    \textbf{Lei Liang}$^\clubsuit$,\\
    \textbf{Wen Zhang}$^\spadesuit$$^\diamondsuit$\footnotemark[1],
    \textbf{Huajun Chen}$^\spadesuit$$^\diamondsuit$\thanks{~~Corresponding authors.}\\
    $^\spadesuit$ Zhejiang University \\
    $^\diamondsuit$ Zhejiang University - Ant Group Joint Laboratory of Knowledge Graph\\
    $^\clubsuit$ Ant Group\\
    \texttt{
    \{zhangyichi2022,zhuo.chen,zhang.wen,huajunsir\}@zju.edu.cn 
    }\\
  \faGithub\,\url{https://github.com/zjukg/SUBARU}
}

\begin{document}
\maketitle
\begin{abstract}
Large language models (LLMs) have demonstrated exceptional performance in text generation within current NLP research. However, the lack of factual accuracy is still a dark cloud hanging over the LLM skyscraper. Structural knowledge prompting (SKP) is a prominent paradigm to integrate external knowledge into LLMs by incorporating structural representations, achieving state-of-the-art results in many knowledge-intensive tasks. However, existing methods often focus on specific problems, \textbf{lacking a comprehensive exploration of the generalization and capability boundaries of SKP}. This paper aims to evaluate and rethink the generalization capability of the SKP paradigm from four perspectives including \textbf{Granularity, Transferability, Scalability, and Universality}. To provide a thorough evaluation, we introduce a novel multi-granular, multi-level benchmark called SUBARU, consisting of 9 different tasks with varying levels of granularity and difficulty. Through extensive experiments, we draw key conclusions regarding the generalization of SKP, offering insights to guide the future development and extension of the SKP paradigm. 
\end{abstract}

\section{Introduction}

Large language models (LLMs) \cite{DBLP:journals/corr/llmsurvey} have sparked a new wave in the natural language processing (NLP) field. By pre-training on massive corpus with billion-scale decoder transformers \cite{transformer}, LLMs achieve exceptional capabilities in text generation, and are widely used in current researches and applications \cite{KnowPAT, LLM4REC, DBLP:journals/corr/mllm-survey}.
\par
\begin{figure}[t]
    \centering
    {\includegraphics[width=\columnwidth]{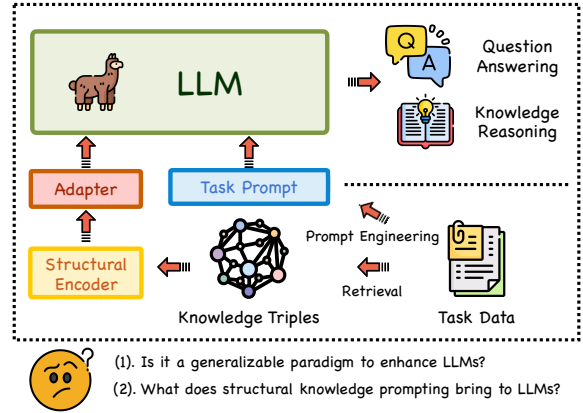}
    }
    \vspace{-16pt}
    \caption{An intuition of the structural knowledge prompting paradigm in current LLM research.}
    \label{figure:introduction}
    \vspace{-16pt}
\end{figure}

 However, the lack of factual accuracy in LLMs \cite{DBLP:journals/corr/Hallucination} remains a significant issue, leading to unreliable and untrustworthy outputs that limit their applications. To address this, external knowledge bases are widely used \cite{RAG} to incorporate reliable knowledge into LLMs for a fact-grounded generation. Among these, knowledge graphs (KGs) \cite{KGSURVEY-TPAMI24} are a specialized form of semi-structured knowledge base, organizing vast amounts of factual knowledge as triples within graph structures. Numerous works \cite{mindmap} propose different KG-oriented methods to incorporate the KGs into the LLMs for specific tasks such as question answering \cite{GNP-AAAI24}, knowledge graph reasoning \cite{KoPA-MM24}. As illustrated in Figure~\ref{figure:introduction}, structural knowledge prompting ({\prompt}) is a widely used paradigm that integrates pre-trained structural information in the KGs to the LLMs with an adapter. The structural information is learned in the structural encoder, while the adapter is an extra neural network to bridge the representation gap.
This approach is consistent with many Multi-modal LLMs (MLLMs) \cite{DBLP:journals/corr/mllm-survey}, where a pre-trained encoder is used to bridge the non-textual information to the textual representation space of the LLMs.
\par However, existing {\prompt} methods typically adapt the paradigm directly to specific tasks with the ready-to-use principle, without thoroughly examining the paradigm itself. This raises several important questions: What makes {\prompt} successful on specific tasks? What level of knowledge granularity does {\prompt} provide to LLMs, and how can it enhance generalization across tasks of varying difficulty? Furthermore, it is crucial to re-assess the full spectrum of {\prompt} methods to better guide the future development of this research field.
\par To fill the gaps in current research, we explore the generalization capability of the {\prompt} paradigm in this paper. We first construct a new \underline{\textbf{S}}tr\underline{\textbf{U}}ctural Prompting \underline{\textbf{B}}enchm\underline{\textbf{A}}rk with \underline{\textbf{R}}easoning and \underline{\textbf{U}}nderstanding tasks ({\benchmark} for short), consisting of 9 tasks with varying granularity and difficulty, which are captured from large-scale encyclopedic knowledge graphs. We design a complete training evaluation protocol to adequately assess the generalization of {\prompt} across four dimensions: \textbf{Granularity, Transferability, Scalability, Universality}. Finally, we conduct extensive experiments with 16 different {\prompt} settings, exploring these four dimensions and drawing key conclusions to explain the success of current {\prompt} methods, while offering insights to guide future developments aimed at enhancing the factual accuracy of LLMs. Our contribution can be summarized as:
\begin{itemize}
    \item We examine the widely used {\prompt} paradigm in current LLM research. Rather than applying it to specific tasks, we provide a systematic evaluation and explore its generalization potential.
    \item We introduce a new benchmark, {\benchmark}, consisting of 9 tasks with varying granularity and difficulty levels, designed to assess the generalization of the {\prompt} paradigm.
    \item We conduct a comprehensive evaluation on the generalization of existing {\prompt} modules from four dimensions, exploring the granularity, transferability, scalability, and universality. We make several interesting conclusions after the explorations.
\end{itemize}
\section{Related Works}
The combination of KGs and LLMs \cite{KoPA-MM24, MKGL, LightRAG, HippoRAG, LLMKG1} is an important topic in nowadays research. In addition to the factual knowledge contained in KGs, many methods try to augment the LLMs with the rich structural information present in the KG to achieve knowledge infusion capabilities. DrugChat \cite{DrugChat-TechRxiv23} and GNP \cite{GNP-AAAI24} employ a graph neural network to extract structural information from the retrieved knowledge subgraph to enhance the question-answering (QA) ability of LLMs. KoPA \cite{KoPA-MM24} incorporates the pre-trained structural knowledge embeddings into LLMs with a project layer to enhance the knowledge graph completion (KGC) ability of LLMs. Their paradigm lies in the use of various structural encoders to extract non-textual features and for enhancing the textual inference capability of LLM, a concept borrowed from multi-modal LLMs. While adaptations are made for specific tasks, there is a lack of in-depth exploration on the rationale of this paradigm. In this paper, we provide a comprehensive evaluation and analysis of its generalization ability.
\begin{figure*}[t]
    \centering
    {\includegraphics[width=\textwidth]{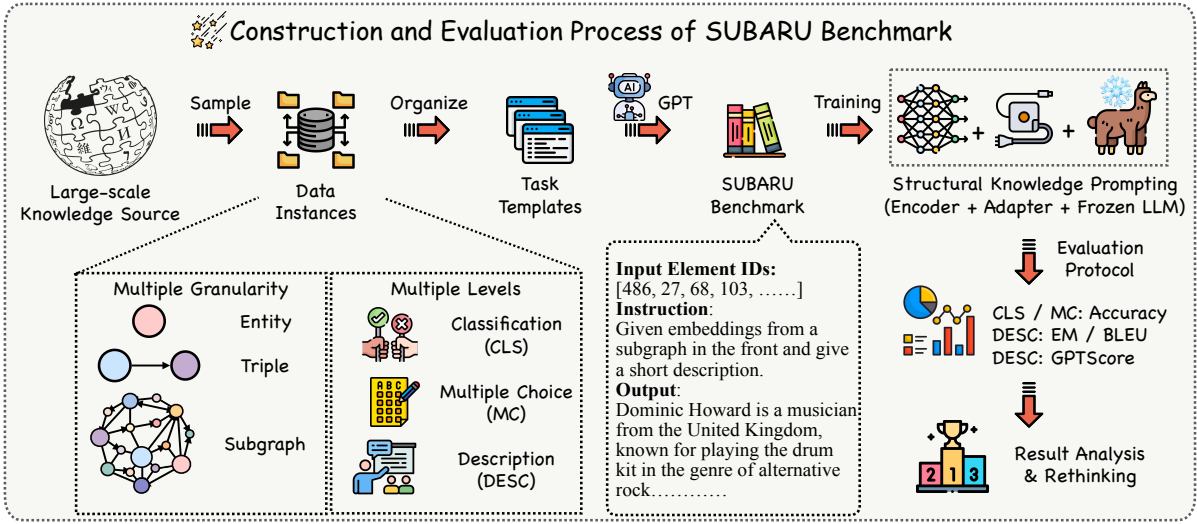}
    }
    \vspace{-16pt}
    \caption{An overview of the construction pipeline of {\benchmark} benchmark and the evaluation process. In {\benchmark}, we construct 9 different tasks with multiple granularity (entity/triple/subgraph) and multiple difficulty levels for comprehensive evaluation of the generalization capability of the structural knowledge prompting paradigm.}
    \label{figure:framework}
    \vspace{-16pt}
\end{figure*}
\section{Preliminary}
\label{sec:preliminary}
In this paper, we focus on the evaluation of the structural knowledge promptings (\prompt) in the LLMs. The LLM is denoted as $\mathcal{M}$. and the general prediction process of the LLM can be denoted as:
\begin{equation}
    \mathcal{A}^*=\max_{\mathcal{A}} P_{\mathcal{M}}(\mathcal{A}|\mathcal{Q})
\end{equation}
where $\mathcal{A}$ is the answer to the question $\mathcal{Q}$, and $\mathcal{A}^*$ is the optimal answer decoded by the LLMs.
\par Many recent works aim to enhance the reasoning ability of LLMs by incorporating structural knowledge. A common approach involves retrieving relevant entities and relations from an external KG, extracting and embedding them as prompt tokens for the LLM. We denote an external KG as $\mathcal{KG}=(\mathcal{E}, \mathcal{R}, \mathcal{T})$, where $\mathcal{E}, \mathcal{R}, \mathcal{T}$ are the entity set, relation set and triple set respectively. A triple $(h, r, t)$ means that there is a relation $r$ between head entity $h$ and tail entity $t$. An entity or a relation would be treated as one basic element $s_i$ in such a {\prompt} process. For a given element $e_i$, the input prompt token can be denoted as:
\begin{equation}
    \mathcal{S}(e_{i})=\mathcal{P}(\mathtt{ENC}(e_i|\mathcal{KG}))
\end{equation}
where $\mathtt{ENC}(e_i|\mathcal{KG})$ is the structural encoder learned self-supervisedly on the given $\mathcal{KG}$ and $\mathcal{P}$ is the adapter for bridging two representation spaces of the structural embeddings and LLMs. Several classic implementations of the adapter $\mathcal{P}$ exist, such as a simple MLP \cite{GNP-AAAI24}, Qformer \cite{BLIP2}, etc. Depending on the specific task, the {\prompt} tokens would be organized as a sequence $\mathcal{S}=(\mathcal{S}_1, \dots, \mathcal{S}_n)$, which can represent a single entity, a triple, or a subgraph. It is further concatenated with the input tokens of the LLMs, resulting in the final prediction process:
\begin{equation}
    \mathcal{A}^*=\max_{\mathcal{A}} P_{\mathcal{M}}(\mathcal{A}|\mathcal{Q}, \mathcal{S})
\end{equation}
\par This basic setting and formulation for {\prompt} are widely used in existing works. However, current approaches often this paradigm directly to specific downstream tasks \textbf{without a deeper exploration of the paradigms themselves}. In this paper, we will dive deep into this problem by conducting a comprehensive evaluation with our proposed benchmark to address this gap.
\section{Our Evaluation Framework}
\label{sec:evaluation-framework}
\subsection{The General Motivation}
\label{sec:motivation}
In the previous section, we introduce the basic paradigm of {\prompt}, which is widely used by current KG-enhanced LLM applications. While these methods have achieved state-of-the-art results in knowledge-intensive tasks such as QA and KGC, the generalization capabilities of the {\prompt} paradigm remain under-explored. In this paper, we will explore the following four research questions (RQ) about the generalization ability of {\prompt}:
\begin{itemize}
    \item \textbf{RQ1. Granularity}: What levels of structural knowledge from KGs can the {\prompt} paradigm integrate to LLMs?
    \item \textbf{RQ2. Transferability}: Is the {\prompt} paradigm transferable across different tasks? Can {\prompt} process new elements haven't seen before?
    \item \textbf{RQ3. Scalability}: Does the {\prompt} paradigm exhibit scaling laws?
    \item \textbf{RQ4. Universality}: Can the {\prompt} paradigm be applied to different LLMs?
\end{itemize}
Existing {\prompt} methods lack the exploration of the above four questions, and no suitable benchmarks exist for such exploration. To facilitate a more thorough investigation, we start by introducing a new benchmark containing various new datasets and tasks to facilitate better nature exploration in the experimental section.

\begin{table}
\centering
\caption{The statistical information of {\benchmark}. We have 3 granularity and 3 levels resulting in 9 tasks.}
\label{tab:benchmark}
\resizebox{0.45\textwidth}{!}{
\begin{tabular}{cc|ccc}
\toprule
\multicolumn{2}{c|}{\textbf{Task}} & \textbf{\# Train} & \textbf{\# Valid} & \textbf{\# Test} \\
\midrule
\multirow{3}{*}{\textbf{Entity (EG)}} & CLS & 32122 & 4016 & 4016 \\
 & MC & 16096 & 2012 & 2013 \\
 & DESC & 16061 & 2008 & 2008 \\
 \midrule
\multirow{3}{*}{\textbf{Triple (TG)}} & CLS & 371168 & 20620 & 20622 \\
 & MC & 185584 & 10310 & 10311 \\
 & DESC & 185584 & 10310 & 10311 \\
 \midrule
\multirow{3}{*}{\textbf{Subgraph (SG)}} & CLS & 29454 & 3998 & 5142 \\
 & MC & 14727 & 1999 & 2571 \\
 & DESC & 7453 & 931 & 939 \\
\bottomrule
\end{tabular}
}
\end{table}

\subsection{The {\benchmark} Benchmark}
To better explore the mentioned four key RQs about the {\prompt} paradigm. We propose the \underline{\textbf{S}}tr\underline{\textbf{U}}ctural Prompting \underline{\textbf{B}}enchm\underline{\textbf{A}}rk with \underline{\textbf{R}}easoning and \underline{\textbf{U}}nderstanding tasks ({\benchmark} for short). In this section, we we briefly introduce {\benchmark}, outlining its general principles and the process of its construction. An overview of the {\benchmark} framework is presented in Figure~\ref{figure:framework}.

\subsubsection{General Principle of {\benchmark}}
Existing {\prompt} applications typically target the specific tasks requiring varying granularity of structural knowledge from KGs. For instance, the QA task needs subgraph-level knowledge, while the KGC task only requires triple-level knowledge. Although these applications have made significant progress using the same paradigm, they \textbf{do not fully capture the capabilities of {\prompt}}. In the {\benchmark} framework, we aim to evaluate the {\prompt} paradigm more comprehensively by designing tasks with different levels of structural knowledge granularity. These include entity-granularity (EG), triple-granularity (TG), and subgraph-granularity tasks (SG), which assess the model's ability to reason and understand entities, triples, and subgraphs from KGs.

\par Additionally, depending on the difficulty of reasoning and comprehension, we introduce three more difficulty levels in our {\benchmark}: binary classification (CLS), multiple choice (MC), and description (DESC). These tasks correspond to the model's ability to perform binary classification, answer multiple-choice questions, or generate detailed descriptions based on the input structural prompts. By combining the three granularity with three levels of difficulty, we create \textbf{9 different tasks} as shown in Figure~\ref{figure:framework}. Next, we describe how our dataset was constructed.

\subsubsection{Construction Process of {\benchmark}}
We present an overview of the {\benchmark} in Table \ref{tab:benchmark}. We employ CoDeX \cite{CoDeX}, a large-scale KG extract from WikiData \cite{DBLP:journals/cacm/wikidata} as our data source. CoDeX contains approximately 110K triples. The construction process involves two key steps:

\noindent\textbf{Instance Sampling.} First, we sample entity/triple/subgraph instances at different granularity from the KG to prepare for different tasks. For the EG task, we sample approximately 20K entities with adequate descriptions with an 8:1:1 split. For the TG task, we employ the split of CoDeX-M triples to build the datasets. For the SG task, we start with the entities selected in the EG task and then randomly sample their 1-hop and 2-hop neighborhoods to construct the subgraphs.
Meanwhile, each task has specific settings. For the CLS task, we treat an entity ID with its real short name as a positive instance. For TG and SG, we consider each triple and subgraph sampled from the existing KG as positive instances. We further generate negative samples maintaining a 1:1 ratio by random perturbing. In the MC task, we sample four choices for each instance: for EG, we predict the entity name, and for TG and SG, we predict the missing entity. The missing entity prediction in TG-MC is similar to the traditional KGC task to predict the missing tail entity in the given query $(h, r, ?)$. For SG, the query provides a subgraph with one core entity missing and ask for the missing entity in the subgraph. For the DESC task, the entity, triple, and subgraph descriptions serve as the target for generation. Entity and triple descriptions are taken directly from the CoDeX dataset, while subgraph descriptions are generated using \textit{GPT-3.5-turbo}. Due to the page limit, we present a more detailed description of the 9 tasks in Appendix \ref{sec:appendix:task_settings}.

\noindent\textbf{Prompt Generation.} After sampling from the CoDeX KG, we create task-specific instances by applying a hand-crafted instruction prompt, $\mathcal{I}_{task}$, for each task, transforming the instances into the text format for further evaluation. Following the existing paradigms, we put the {\prompt} in the front of the input sequence to inform the LLMs with structural information from KGs. To objectively assess the model's ability to utilize these {\prompt}s, we remove the important textual information of the relevant elements in the instruction template, \textbf{allowing the model to complete the tasks using mainly the {\prompt}s rather than the texts to assess the utilization of the {\prompt}s.}. We present a general prompt template used in our evaluation in Figure~\ref{box:all_tempelate}. Here we present a general prompt template during our evaluation. We present the detailed prompt templates and data samples in Appendix \ref{sec:appendix:prompt_template}.
\begin{figure}
\begin{tcolorbox}[colback=green!10!white, colframe=green!60!black, title=Task Prompt Template in {\benchmark}]
\label{box:entity_caption}
\textbf{Input:} <Structural Knowledge Promptings $\mathcal{S}$>, <Task-specific Prompt $Q_{task}$> \\
\textbf{Output:} Task-specific Answers $\mathcal{A}$
\end{tcolorbox}
\caption{A general prompt template for all tasks.}
\label{box:all_tempelate}
\vspace{-16pt}
\end{figure}

\begin{table*}[]
\centering
\caption{The main experiment results on the 9 tasks of {\benchmark} benchmark. We colored the top-3 results under each task with a different \textcolor{green2}{green} color from shallow to deep.}
\label{table:main}
\resizebox{\textwidth}{!}{
\begin{tabular}{cc|cccccccccc}
    \toprule
    \multicolumn{2}{c|}{} & \multicolumn{3}{c}{\textbf{Entity Granularity}} & \multicolumn{3}{c}{\textbf{Triple Granularity}} & \multicolumn{4}{c}{\textbf{Subgraph Granularity}} \\
    \multicolumn{2}{c|}{\multirow{-2}{*}{\textbf{Method}}} & \textbf{L-1 (Acc)} & \textbf{L-2 (Acc)} & \textbf{L-3  (EM)} & \textbf{L-1 (Acc)} & \textbf{L-2 (Acc)} & \textbf{L-3  (EM)} & \textbf{L-1 (Acc)} & \textbf{L-2 (Acc)} & \textbf{L-3  (B-4)} & \textbf{L-3 (GPT)} \\
    \midrule
    \multicolumn{2}{c|}{\textbf{Random Choice}} & 50.00 & 25.00 & - & 50.00 & 25.00 & - & 50.00 & 25.00 & - & - \\
    \midrule
         & \textbf{TransE} & 55.85 & 39.49 & 0.00 & 55.41 & 87.42 & 2.77 & 82.43 & 57.95 & 9.85 & 14.97 \\
         & \textbf{DistMult} & 52.61 & {34.02} & 0.00 & 47.23 & 89.60 & 21.21 & 87.10 & 78.99 & 4.32 & 35.42 \\
         & \textbf{RotatE} & 57.34 & 51.16 & 0.00 & 55.59 & 66.77 & {6.88} & 70.12 & 62.54 & 2.12 & 22.20 \\
    \multirow{-4}{*}{\textbf{FC}} & \textbf{R-GCN} & 52.44 & 41.97 & 0.00 & 52.53 & 90.50 & 27.90 & {86.75} & 54.49 & {5.64} & 19.57 \\
    \midrule
         & \textbf{TransE} & \UR 84.76 & \UR 91.01 & 0.00 & 53.92 & 86.51 & 41.83 & \SSR 91.71 & 89.34 & 12.67 & 45.61 \\
         & \textbf{DistMult} & 57.71 & 61.84 & 0.00 & \SR 55.64 & 93.53 & \SSR 97.91 & 65.46 & \SR 90.35 & \SSR 26.72 & \UR 55.56 \\
         & \textbf{RotatE} & \SSR{85.43} & 54.94 & 0.00 & 53.71 & 88.21 & \SR 83.74 & \SR 89.14 & \UR 90.43 & \UR 24.74 & \SSR 55.80 \\
        \multirow{-4}{*}{\textbf{MLP}} & \textbf{R-GCN} & 66.85 & 44.46 & 0.00 & 53.68 & 90.45 & \UR 94.59 & 76.70 & \SSR 91.05 & 16.08 & \SR 49.25 \\
    \midrule    
         & \textbf{TransE} & 58.66 & {38.89} & 0.00 & \SSR 65.24 & 92.07 & 19.72 & 88.46 & 80.32 & 19.50 & 46.93 \\
         & \textbf{DistMult} & 55.35 & 27.37 & 0.00 & 54.49 & 92.92 & {8.38} & 86.98 & 81.60 & 9.75 & 35.57 \\
         & \textbf{RotatE} & 56.47 & 29.30 & 0.00 & \UR 59.47 & 89.09 & {7.32} & \UR 90.31 & 88.44 & \SR 19.68 & 47.05 \\
    \multirow{-4}{*}{\textbf{MoE}} & \textbf{R-GCN} & 54.23 & 38.15 & 0.00 & 53.59 & 92.09 & 27.46 & 54.90 & 78.91 & 4.04 & 20.39 \\
    \midrule
         & \textbf{TransE} & 59.48 & \SSR 92.50 & 0.00 & 54.77 & \SSR 94.42 & 38.90 & 78.97 & 27.14 & 11.32 & 41.19 \\
         & \textbf{DistMult} & 75.34 & 60.40 & 0.00 & {52.95} & \UR 94.11 & 17.77 & 78.59 & 37.53 & 9.25 & 32.74 \\
         & \textbf{RotatE} & \SR {82.96} & \SR 79.50 & 0.00 & 50.84 & \SR 94.02 & 6.23 & 80.35 & 26.33 & 14.43 & 42.41 \\
    \multirow{-4}{*}{\textbf{Q-former}} & \textbf{R-GCN} & 81.77 & 41.11 & 0.00 & 51.18 & 93.94 & 16.23 & 73.60 & 27.42 & 4.73 & 21.82 \\
    \bottomrule
\end{tabular}
}
\end{table*}
\subsubsection{The Evaluation Process of {\benchmark}}
In the following experiments section, we provide a comprehensive evaluation of the four generalization properties of the {\prompt} paradigm using the {\benchmark}. As {\prompt} is an external module added to knowledge-intensive task adaption, it must be trained on the training set before its performance can be evaluated on the test set. The training process, based on classic next-word prediction, is defined as:
\begin{equation}
    \mathcal{L}_{{\prompt}}=-\log P_{\mathcal{M}}(\mathcal{A}|\mathcal{Q}_{task},\mathcal{S})
\end{equation}
where $\mathcal{Q}, \mathcal{A}$ is an question-answering pair from the training data. $\mathcal{S}$ is the corresponding structural prompt for the given question, which can be an entity, a triple, or a sequential subgraph. During training, the LLM $\mathcal{M}$ is frozen, and the adapter $\mathcal{P}$ is trained to bridge the pre-trained structural encoder $\mathtt{ENC}()$ and the LLM. Meanwhile, for the four RQs, we will present the detailed evaluation protocols and implementation in the next section.

\section{Evaluation Results}
\label{sec:experiments}
In this section, we first introduce the experimental setup, including implementation details and the evaluation protocol. Then, we present the results of the experiments to explore the four significant RQs (mentioned in Section \ref{sec:motivation}) about the Granularity, Transferability, Scalability, and Universality of the {\prompt} paradigm. We further provide some intuitive cases to analysis the competency boundary of {\prompt}.

\subsection{Experimental Setup}
\label{sec:experimental-setup}
\noindent\textbf{Implementation Details.} TransE \cite{DBLP:conf/nips/TransE}, DistMult \cite{DBLP:journals/corr/DistMult}, RotatE \cite{DBLP:conf/iclr/RotatE}, and R-GCN \cite{DBLP:conf/esws/RGCN}. We implement the structural encoders based on NeuralKG \cite{DBLP:conf/sigir/NeuralKG}. Among these structural encoders, R-GCN is a graph neural network method and others are classic KG embedding methods. For the adapter $\mathcal{P}$, we choose 4 mainstream architectures used by recent works, including single \textbf{fully-connected layer (FC)} \cite{KoPA-MM24}, \textbf{multi-layer perceptron (MLP)} \cite{DrugChat-TechRxiv23}, \textbf{MoE} \cite{XRec}, and \textbf{Qformer} \cite{BLIP2}. For LLMs, we mainly employ Llama3-8B-Instruct \cite{llama3} as the backbone model for experiments. We also evaluate the performance of other LLMs \cite{DBLP:journals/corr/llama2} on the {\benchmark} in further exploration. All the experiments are conducted on a Linux server with NVIDIA A800 GPUs. We set the LLMs in FP16 precision and optimized the {\prompt} with AdamW \cite{Adamw-ICLR22} optimizer. For detailed backbone selection, hyper-parameter settings, and training efficiency of all the tasks, we present a summary in Appendix \ref{sec:appendix:details}.

\noindent\textbf{Evaluation Protocol.} For the CLS and MC tasks, we use the accuracy (ACC) for evaluation. For the description tasks, we use different metrics for different granularity levels. For the entity-level and triple-level tasks, we use the exact match (EM) rate as the evaluation metric, which needs the LLMs to generate the exact entity name and triple information.  For the subgraph-level description, we use BLEU-4 (B-4) \cite{DBLP:conf/acl/BLEU} and GPTScore as the evaluation metrics. BLEU is a traditional metric for text generation evaluation. GPTScore follows the LLM-as-a-judge paradigm \cite{llmasajudge} and employs \textit{GPT-3.5-tubor} as the judger to score the generated descriptions against the golden answer. A more detailed introduction of our evaluation protocol and the prompt template for GPT scoring can be found in Appendix \ref{sec:appendix:evaluation}.

\subsection{Multi-Granularity Knowledge Evaluation (RQ1)}

The main evaluation results are presented in Table \ref{table:main}. Based on these results, we make the following observations regarding the granularity of structural knowledge learned by the models:

\noindent\textbf{Observation 1. Simple MLP is surprisingly effective.} Despite recent efforts to use complex adapters for {\prompt}, the simple MLP architecture achieves the best performance on most tasks in {\benchmark}. As presented in Table~\ref{table:main}, the MLP-based results dominate in most of the colored cells compared to other adapters. Complex {\prompt} architectures like Qformer and MoE don't perform well on certain tasks like SG.

\noindent\textbf{Observation 2. {\prompt} excels in coarse-grained reasoning tasks.} The three granularities in {\benchmark} correspond to progressively coarser reasoning tasks (MC). For EG, LLMs must precisely understand the input {\prompt} to make the correct choice. However, for TG and SG, the MC task becomes more coarse-grained, with the LLM only needing to grasp the correlation between the input {\prompt} and the options, as the {\prompt} provides more semantic richness and auxiliary information. We can observe that {\prompt} performs well in the MC tasks of TG and SG than EG. This suggests that SKP demonstrates some coarse-grained reasoning ability, but lacks sufficient fine-grained understanding for EG.

\noindent\textbf{Observation 3. {\prompt} struggles to understand new entities accurately, failing in the EG DESC task.} As we mentioned before, the smallest elements in {\prompt} are entities or relations. Therefore, in the EG task, all the tasks require LLMs to understand the unseen entities during training and make predictions or descriptions. We observe that {\prompt} performs poorly in the EG task, especially in the DESC task. This suggests that {\prompt} struggles to accurately understand new entities and \textbf{lacks inductive reasoning ability at EG}. This is because the current {\prompt} modeling approach is still relatively lacking in extrapolation capabilities. The encoder's characterization ability is inadequate for effective extrapolation when bridged with the LLM, highlighting a gap between {\prompt} and classical MLLM in terms of generalization.

\par Based on these observations, we conclude that current {\prompt} methods are \textbf{not perfect across all granularities and have their limitations}. However, mainstream {\prompt} applications typically focus on triple or subgraph reasoning tasks, where these methods excel. The format of these downstream tasks closely resembles the TG/SG MC tasks. Additionally, to better understand the description ability of {\prompt} models, we further conduct a case study in Figure~\ref{sec:case}.
\subsection{Transferability Evaluation (RQ2)}
\begin{figure}[t]
    \centering
    {\includegraphics[width=\columnwidth]{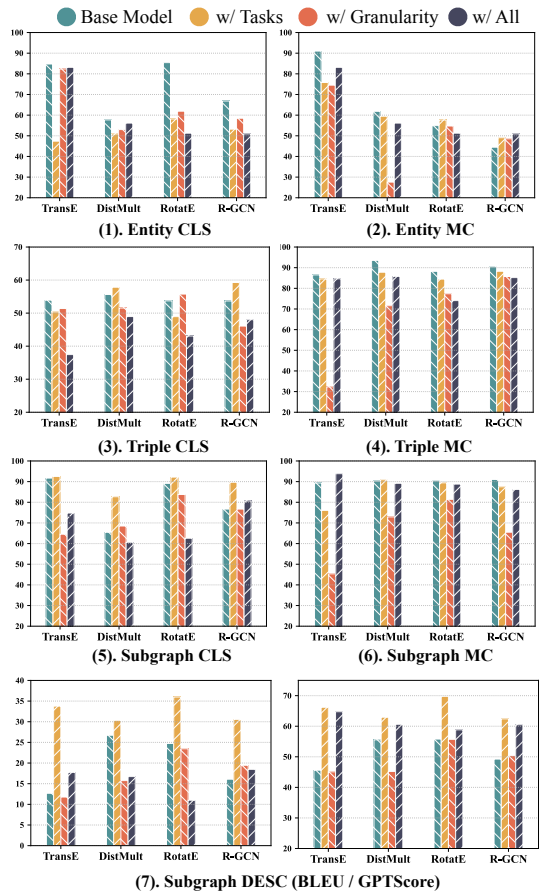}
    }
    \vspace{-16pt}
    \caption{The transferability experiments among different granularities and levels with MLP adapter.}
    \label{figure:transfer}
    \vspace{-16pt}
\end{figure}
To further validate the transferability of {\prompt} methods, we conduct an additional evaluation to answer the following two sub-issues: (1). Can {\prompt} learn positive transfer from the tasks in different granularities and levels? (2). How well does {\prompt} handle new entities under different scenarios? These issues relate to the \textbf{transferability of {\prompt} across tasks and elements}.
\subsubsection{Transferability among Tasks}
\noindent\textbf{Settings.} We conduct four sets of experiments to explore this issue by training {\prompt} models with the dataset from the single task (Base), all tasks in the same level, same granularity, and whole benchmark (w/ Tasks, w/ Granularity, w/ All). The model’s performance is then evaluated on each task to investigate whether it benefits from knowledge transfer across tasks.

\noindent\textbf{Analysis.} The results are shown in Figure~\ref{figure:transfer} reveal that {\prompt} does not exhibit strong transferability on CLS and MC tasks. In most cases,training on tasks across different granularities or difficulty levels does not yield a significant improvement in model performance on the target task. However, {\prompt} models perform better on the DESC task when trained with additional data, likely due to the nature of the task. SG DESC is a relatively coarse-grained generation task that can benefit from extra training data containing more structural information.
\par Overall, this experiment suggest that the current {\prompt} architecture faces challenges in transferability.

\begin{figure}[t]
    \centering
    {\includegraphics[width=\columnwidth]{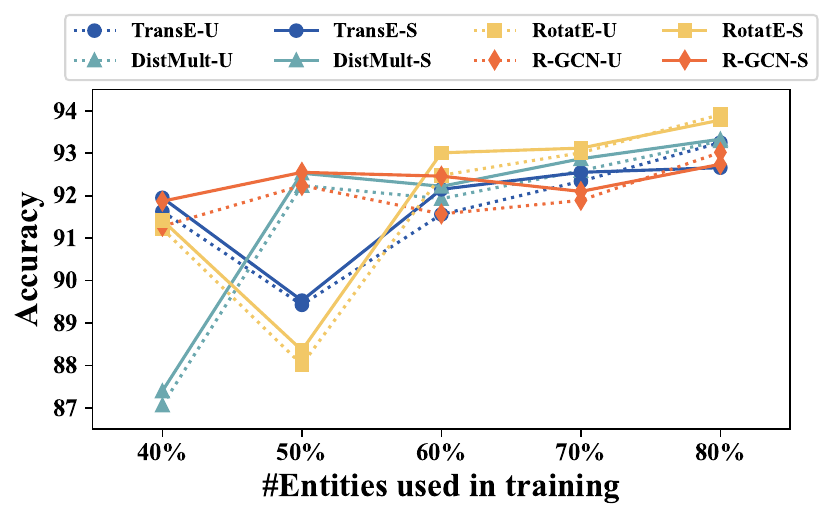}
    }
    \vspace{-16pt}
    \caption{The inductive transfer experiments.}
    \label{figure:inductive}
    \vspace{-16pt}
\end{figure}

\subsubsection{Transferability in New Elements}
To explore {\prompt}'s ability to handle new entities, we conduct experiments in an inductive transfer scenario for TG tasks. As we mentioned before, the entity is the basic element in {\prompt}, making EG tasks naturally inductive. While SG Task will inevitably have overlapping elements, TG is the best scenario for inductive experiments.

\noindent\textbf{Settings.} We re-split the datasets by preserving a certain ratio (\%) of entities in the training set. In the test set, there will be some unseen entities. This allows us to further divide the test triples into two categories: triples with unseen entities (U) and those without (S). We evaluate the model's performance on these two subsets separately.

\noindent\textbf{Analysis.} The results in Figure~\ref{figure:inductive} suggest that {\prompt} performs well with new entities in the TG MC task. Specifically, the performance on the unseen triples is nearly identical to that on the seen triples. Besides, training on more entities can improve the model's ability to inductive reasoning. This raises an interesting question: why does {\prompt} perform better in terms of transferability on TG tasks compared to EG tasks? We believe this is due to a combination of task difficulty and granularity. The {\prompt}s in TG and SG tasks provide more structural context, which reduces the difficulty for the model in understanding the key elements, making it easier to generate correct predictions.
\begin{table*}[t]
\caption{A case study for subgraph description task. We colored the key words of the sentence.}
\label{table:case}
\vspace{-8pt}
\centering
\resizebox{\textwidth}{!}{
\begin{tabular}{l}
\toprule
\midrule
\begin{tabular}[c]{@{}l@{}}\textbf{(Golden Answer)} \textbf{Franz Grashof} was a \textcolor{red}{university teacher} in the field of \textcolor{purple}{mechanics}, who \textcolor{orange}{spoke, wrote, or signed in} \textcolor{blue}{German}. \\ He was \textcolor[rgb]{0,0.4,0}{buried at Hauptfriedhof Karlsruhe}.\end{tabular}                                                                                    \\
\midrule
\begin{tabular}[c]{@{}l@{}}\textbf{(Good Prediction)} \textbf{Friedrich Ernst} Petersen was a \textcolor{blue}{German} engineer who worked \textcolor{purple}{in the field of engineering}. He was \textcolor{red}{employed at the} \\ \textcolor{red}{University} of Göttingen, where he was also educated. Friedrich Ernst Petersen \textcolor{orange}{spoke, wrote, or signed in} \textcolor{blue}{German}. \textbf{(GPTScore: 90)}\end{tabular}\\
\midrule
\begin{tabular}[c]{@{}l@{}}\textbf{(Neutral Prediction)} \textbf{Friedrich August} von Hayek was a \textcolor{red}{university teacher} who specialized in the field of economics. He was fluent  \\ in \textcolor{blue}{German}, which he \textcolor{orange}{spoke, wrote, and signed}. After his death, he \textcolor[rgb]{0,0.4,0}{was buried in} the cemetery of the village of Zermatt. \textbf{(GPTScore: 60)}\end{tabular}  \\

\midrule
\begin{tabular}[c]{@{}l@{}}\textbf{(Bad Prediction)} \textbf{Karl-Heinz Rummeny}, a notable alpine skier, was born in Garmisch-Partenkirchen. \\ He was also a member of the \textcolor{blue}{German} Alpine Club and was also a member of the German Ski Association. \textbf{(GPTScore: 25)}\end{tabular}  \\

\midrule
\bottomrule
\end{tabular}
}
\end{table*}
\subsection{Scalability Evaluation (RQ3)}
\begin{figure}[t]
    \centering
    {\includegraphics[width=\columnwidth]{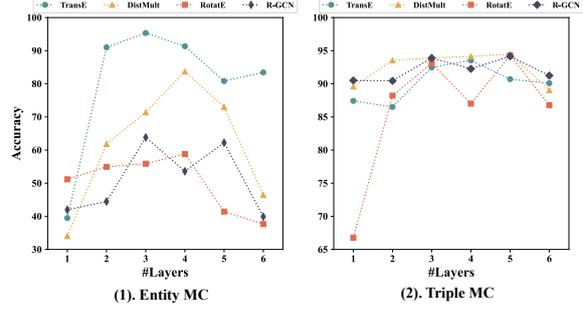}
    }
    \vspace{-16pt}
    \caption{The scalability experiments on the entity / triple MC tasks. We use MLP as the adapter.}
    \label{figure:scala}
    \vspace{-16pt}
\end{figure}

To verify the scalability, we conducted the following experiment.

\noindent\textbf{Settings.} We dedicate research to the MLP adapter, constantly deepening its layers from 1 to 6 for experiments on the {\benchmark} benchmark.

\noindent\textbf{Analysis.} The results presented in Figure~\ref{figure:scala} reveal that adapters in {\prompt} models undergo an upward and then downward change in performance. An MLP with 3-4 layers is strong enough to achieve state-of-the-art performance. This law is confirmed on different structural encoders. Of course, the scalability of the adapter is also related to the amount of data required for training, which is currently limited, and the scalability at larger data volumes needs to be further explored.

\subsection{Universality Evaluation (RQ4)}
\begin{figure}[t]
    \centering
    {\includegraphics[width=0.85\columnwidth]{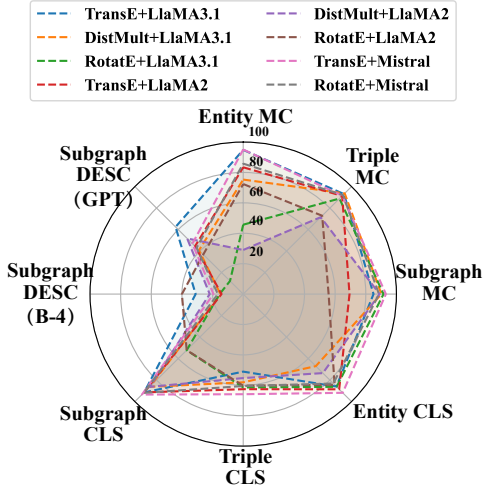}
    }
    \vspace{-6pt}
    \caption{Experiments on more different LLMs.}
    \label{figure:universe}
    \vspace{-12pt}
\end{figure}
\noindent\textbf{Settings.} To further evaluate the universality of {\prompt}, we conduct experiments with more different LLM backbones such as Llama2-7B-chat \cite{DBLP:journals/corr/llama2}, Llama3.1-8B-Instruct \cite{llama3}, and Mistral-7B \cite{mistral}. We report the results of them on the MLP adapter.

\noindent\textbf{Analysis.} As shown in Figure~\ref{figure:universe}, we can find that {\prompt} is a universal framework that can be adapted in various decoder LLMs. The performance variation across different backbones on the same task is minimal, and a consistent trend is observed across models. However, there is a slight performance drop when combining certain backbones, likely due to differences in the LLM architectures. More additional results about LlaMA2 and LlaMA3.1 are presented in Appendix \ref{sec:appendix:llama}.
\subsection{Case Study and Further Analysis}
\label{sec:case}
In the three difficulty levels we designed in {\benchmark} benchmark, the CLS and MC tasks provide clear answers and quantitative metrics, allowing for precise comparisons of model performance. However, for the task of subgraph DESC, assessing the quality of generated text is more subjective. Therefore, we conduct case studies in this section to analyze the ability of the {\prompt} model to describe the subgraph structures. \textbf{The goal of this case study is not to compare the differences in the performance of different {\prompt} models, but to identify the commonalities that exist in descriptions.} As shown in Figure~\ref{table:case}, we present a simple case with a golden answer and several predictions from different {\prompt} models, which would be a personal description. We can observe the following two points: 
\par (1). None of the {\prompt} models were able to accurately identify the central entity, highlighting the \textbf{inability to include particularly precise and personalized information} in the {\prompt}s. This also explains why all the {\prompt} models fail in the EG DESC task in Table~\ref{table:main}, which requires precise entity identification.
\par (2). \textbf{The {\prompt} models demonstrate an understanding of some coarse-grained entities and relations} in the input {\prompt}, capturing their connections and reflecting semantic understanding in the generated text.  A good prediction can understand more hidden information encoded in {\prompt} such as profession, major, nationality, and skills. 
\par Combining these insights, we can conclude that {\prompt} can inform LLMs coarse-grained information for understanding some subgraph structures roughly, , but struggles with detailed information like specific names, places, or specialized terms. While {\prompt} excels at recognizing broad knowledge such as entity attributes, it lacks the cognitive ability to handle finer details. As text generation and deep-level understanding are key capabilities for LLMs, we think that \textbf{future improvements to {\prompt} should focus on activating more precise and detailed information through additional prompt tokens}.

\section{Conclusion}
In this paper, we investigate a popular paradigm {\prompt} which aims to integrate external structural knowledge into LLMs. We conduct a thorough evaluation of its generalization capabilities using a new benchmark, {\benchmark}, which encompasses multiple levels of granularity and difficulty. We detail the construction process and evaluation protocol of {\benchmark}. After conducting sufficient experiments in four perspectives, we draw several insightful conclusions. Our findings suggest that {\prompt} effectively provides LLMs with coarse-grained information across different granularities and task types. However, \textbf{achieving fine-grained, precise factual awareness remains a significant challenge}. This evaluation will guide the future development of the {\prompt} to incorporate multiple granularity structural knowledge and task-solving abilities into LLMs.

\section*{Limitations}
In this paper, we make a deep exploration of the generalization of the structural knowledge prompting paradigm. Our work has the following three limitations:
\paragraph{The scale of the {\benchmark} benchmark.} The benchmark constructed by us has some limitations in terms of scale. {\benchmark} does not consist of million-scale training and evaluation data, which limits the exploration of the scalability.

\paragraph{The exploration on larger LLMs.} Due to the limited computational resources, we mainly conduct experiments on LLMs with 7B-8B parameters. Though most of the {\prompt} works are based on LLMs with the same scale, our exploration lacks results on larger LLMs such as 13B and 70B LlaMA.

\paragraph{Lack of explanation of internal mechanisms.} We mainly evaluate the {\prompt} paradigm by the tasks and metrics defined by the {\benchmark} benchmark, lacking further exploration of the internal mechanisms in LLMs, such as the layer-wise attention weights analysis.

We will continue to solve these limitations.

\section*{Ethical Considerations}
In this paper, all of our research and experiments are conducted on publicly available open-source datasets and models. We construct our evaluation benchmark from open-source data and we will release them for open research. Therefore, there is no ethical consideration in this paper.
\bibliography{custom}

\begin{thebibliography}{33}
\providecommand{\natexlab}[1]{#1}

\bibitem[{Bordes et~al.(2013)Bordes, Usunier, Garc{\'{\i}}a{-}Dur{\'{a}}n, Weston, and Yakhnenko}]{DBLP:conf/nips/TransE}
Antoine Bordes, Nicolas Usunier, Alberto Garc{\'{\i}}a{-}Dur{\'{a}}n, Jason Weston, and Oksana Yakhnenko. 2013.
\newblock Translating embeddings for modeling multi-relational data.
\newblock In \emph{{NIPS}}, pages 2787--2795.

\bibitem[{Chen et~al.(2024)Chen, Tan, Zhang, Yang, Sheng, Zhang, Wang, and Chua}]{LLM4REC}
Yuxin Chen, Junfei Tan, An~Zhang, Zhengyi Yang, Leheng Sheng, Enzhi Zhang, Xiang Wang, and Tat{-}Seng Chua. 2024.
\newblock On softmax direct preference optimization for recommendation.
\newblock \emph{CoRR}, abs/2406.09215.

\bibitem[{Dubey et~al.(2024)Dubey, Jauhri, Pandey, Kadian, Al-Dahle, Letman, Mathur, Schelten, Yang, Fan et~al.}]{llama3}
Abhimanyu Dubey, Abhinav Jauhri, Abhinav Pandey, Abhishek Kadian, Ahmad Al-Dahle, Aiesha Letman, Akhil Mathur, Alan Schelten, Amy Yang, Angela Fan, et~al. 2024.
\newblock The llama 3 herd of models.
\newblock \emph{arXiv preprint arXiv:2407.21783}.

\bibitem[{Gao et~al.(2023)Gao, Xiong, Gao, Jia, Pan, Bi, Dai, Sun, Guo, Wang, and Wang}]{RAG}
Yunfan Gao, Yun Xiong, Xinyu Gao, Kangxiang Jia, Jinliu Pan, Yuxi Bi, Yi~Dai, Jiawei Sun, Qianyu Guo, Meng Wang, and Haofen Wang. 2023.
\newblock Retrieval-augmented generation for large language models: {A} survey.
\newblock \emph{CoRR}, abs/2312.10997.

\bibitem[{Glorot et~al.(2011)Glorot, Bordes, and Bengio}]{relu}
Xavier Glorot, Antoine Bordes, and Yoshua Bengio. 2011.
\newblock Deep sparse rectifier neural networks.
\newblock In \emph{{AISTATS}}, volume~15 of \emph{{JMLR} Proceedings}, pages 315--323. JMLR.org.

\bibitem[{Guo et~al.(2024{\natexlab{a}})Guo, Bo, Chen, Zhang, Chen, Lan, Sun, Zhang, Luo, Li, Zhang, Zhang, and Chen}]{MKGL}
Lingbing Guo, Zhongpu Bo, Zhuo Chen, Yichi Zhang, Jiaoyan Chen, Yarong Lan, Mengshu Sun, Zhiqiang Zhang, Yangyifei Luo, Qian Li, Qiang Zhang, Wen Zhang, and Huajun Chen. 2024{\natexlab{a}}.
\newblock Mkgl: Mastery of a three-word language.

\bibitem[{Guo et~al.(2024{\natexlab{b}})Guo, Xia, Yu, Ao, and Huang}]{LightRAG}
Zirui Guo, Lianghao Xia, Yanhua Yu, Tu~Ao, and Chao Huang. 2024{\natexlab{b}}.
\newblock Lightrag: Simple and fast retrieval-augmented generation.
\newblock \emph{CoRR}, abs/2410.05779.

\bibitem[{Guti{\'{e}}rrez et~al.(2024)Guti{\'{e}}rrez, Shu, Gu, Yasunaga, and Su}]{HippoRAG}
Bernal~Jim{\'{e}}nez Guti{\'{e}}rrez, Yiheng Shu, Yu~Gu, Michihiro Yasunaga, and Yu~Su. 2024.
\newblock Hipporag: Neurobiologically inspired long-term memory for large language models.
\newblock \emph{CoRR}, abs/2405.14831.

\bibitem[{Jiang et~al.(2023)Jiang, Sablayrolles, Mensch, Bamford, Chaplot, Casas, Bressand, Lengyel, Lample, Saulnier et~al.}]{mistral}
Albert~Q Jiang, Alexandre Sablayrolles, Arthur Mensch, Chris Bamford, Devendra~Singh Chaplot, Diego de~las Casas, Florian Bressand, Gianna Lengyel, Guillaume Lample, Lucile Saulnier, et~al. 2023.
\newblock Mistral 7b.
\newblock \emph{arXiv preprint arXiv:2310.06825}.

\bibitem[{Li et~al.(2024)Li, Jiang, Huang, Beigi, Zhao, Tan, Bhattacharjee, Jiang, Chen, Wu, Shu, Cheng, and Liu}]{llmasajudge}
Dawei Li, Bohan Jiang, Liangjie Huang, Alimohammad Beigi, Chengshuai Zhao, Zhen Tan, Amrita Bhattacharjee, Yuxuan Jiang, Canyu Chen, Tianhao Wu, Kai Shu, Lu~Cheng, and Huan Liu. 2024.
\newblock From generation to judgment: Opportunities and challenges of llm-as-a-judge.
\newblock \emph{arXiv preprint arXiv: 2411.16594}.

\bibitem[{Li et~al.(2023)Li, Li, Savarese, and Hoi}]{BLIP2}
Junnan Li, Dongxu Li, Silvio Savarese, and Steven C.~H. Hoi. 2023.
\newblock {BLIP-2:} bootstrapping language-image pre-training with frozen image encoders and large language models.
\newblock In \emph{{ICML}}, volume 202 of \emph{Proceedings of Machine Learning Research}, pages 19730--19742. {PMLR}.

\bibitem[{Liang et~al.(2024)Liang, Meng, Liu, Liu, Tu, Wang, Zhou, Liu, Sun, and He}]{KGSURVEY-TPAMI24}
Ke~Liang, Lingyuan Meng, Meng Liu, Yue Liu, Wenxuan Tu, Siwei Wang, Sihang Zhou, Xinwang Liu, Fuchun Sun, and Kunlun He. 2024.
\newblock A survey of knowledge graph reasoning on graph types: Static, dynamic, and multi-modal.
\newblock \emph{IEEE Transactions on Pattern Analysis and Machine Intelligence}.

\bibitem[{Liang et~al.(2023)Liang, Zhang, Zhang, and Xie}]{DrugChat-TechRxiv23}
Youwei Liang, Ruiyi Zhang, li~Zhang, and Pengtao Xie. 2023.
\newblock Drugchat: Towards enabling chatgpt-like capabilities on drug molecule graphs.
\newblock \emph{TechRxiv}.

\bibitem[{Loshchilov and Hutter(2019)}]{Adamw-ICLR22}
Ilya Loshchilov and Frank Hutter. 2019.
\newblock Decoupled weight decay regularization.
\newblock In \emph{{ICLR} (Poster)}. OpenReview.net.

\bibitem[{Lyu et~al.(2024)Lyu, Yan, Wang, Shi, Yin, Ren, Chen, de~Rijke, and Ren}]{LLMKG1}
Yougang Lyu, Lingyong Yan, Shuaiqiang Wang, Haibo Shi, Dawei Yin, Pengjie Ren, Zhumin Chen, Maarten de~Rijke, and Zhaochun Ren. 2024.
\newblock Knowtuning: Knowledge-aware fine-tuning for large language models.
\newblock In \emph{{EMNLP}}, pages 14535--14556. Association for Computational Linguistics.

\bibitem[{Ma et~al.(2024)Ma, Ren, and Huang}]{XRec}
Qiyao Ma, Xubin Ren, and Chao Huang. 2024.
\newblock Xrec: Large language models for explainable recommendation.
\newblock In \emph{{EMNLP} (Findings)}, pages 391--402. Association for Computational Linguistics.

\bibitem[{Papineni et~al.(2002)Papineni, Roukos, Ward, and Zhu}]{DBLP:conf/acl/BLEU}
Kishore Papineni, Salim Roukos, Todd Ward, and Wei{-}Jing Zhu. 2002.
\newblock Bleu: a method for automatic evaluation of machine translation.
\newblock In \emph{{ACL}}, pages 311--318. {ACL}.

\bibitem[{Paszke et~al.(2019)Paszke, Gross, Massa, Lerer, Bradbury, Chanan, Killeen, Lin, Gimelshein, Antiga et~al.}]{pytorch}
Adam Paszke, Sam Gross, Francisco Massa, Adam Lerer, James Bradbury, Gregory Chanan, Trevor Killeen, Zeming Lin, Natalia Gimelshein, Luca Antiga, et~al. 2019.
\newblock Pytorch: An imperative style, high-performance deep learning library.
\newblock \emph{Advances in neural information processing systems}, 32.

\bibitem[{Safavi and Koutra(2020)}]{CoDeX}
Tara Safavi and Danai Koutra. 2020.
\newblock Codex: {A} comprehensive knowledge graph completion benchmark.
\newblock In \emph{{EMNLP} {(1)}}, pages 8328--8350. Association for Computational Linguistics.

\bibitem[{Schlichtkrull et~al.(2018)Schlichtkrull, Kipf, Bloem, van~den Berg, Titov, and Welling}]{DBLP:conf/esws/RGCN}
Michael~Sejr Schlichtkrull, Thomas~N. Kipf, Peter Bloem, Rianne van~den Berg, Ivan Titov, and Max Welling. 2018.
\newblock Modeling relational data with graph convolutional networks.
\newblock In \emph{{ESWC}}, volume 10843 of \emph{Lecture Notes in Computer Science}, pages 593--607. Springer.

\bibitem[{Sun et~al.(2019)Sun, Deng, Nie, and Tang}]{DBLP:conf/iclr/RotatE}
Zhiqing Sun, Zhi{-}Hong Deng, Jian{-}Yun Nie, and Jian Tang. 2019.
\newblock Rotate: Knowledge graph embedding by relational rotation in complex space.
\newblock In \emph{{ICLR} (Poster)}. OpenReview.net.

\bibitem[{Tian et~al.(2024)Tian, Song, Wang, Wang, Hu, Wang, Chawla, and Xu}]{GNP-AAAI24}
Yijun Tian, Huan Song, Zichen Wang, Haozhu Wang, Ziqing Hu, Fang Wang, Nitesh~V. Chawla, and Panpan Xu. 2024.
\newblock Graph neural prompting with large language models.
\newblock In \emph{{AAAI}}, pages 19080--19088. {AAAI} Press.

\bibitem[{Touvron et~al.(2023)Touvron, Martin, Stone, Albert, Almahairi, Babaei, Bashlykov, Batra, Bhargava, Bhosale, Bikel, Blecher, Canton{-}Ferrer, Chen, Cucurull, Esiobu, Fernandes, Fu, Fu, Fuller, Gao, Goswami, Goyal, Hartshorn, Hosseini, Hou, Inan, Kardas, Kerkez, Khabsa, Kloumann, Korenev, Koura, Lachaux, Lavril, Lee, Liskovich, Lu, Mao, Martinet, Mihaylov, Mishra, Molybog, Nie, Poulton, Reizenstein, Rungta, Saladi, Schelten, Silva, Smith, Subramanian, Tan, Tang, Taylor, Williams, Kuan, Xu, Yan, Zarov, Zhang, Fan, Kambadur, Narang, Rodriguez, Stojnic, Edunov, and Scialom}]{DBLP:journals/corr/llama2}
Hugo Touvron, Louis Martin, Kevin Stone, Peter Albert, Amjad Almahairi, Yasmine Babaei, Nikolay Bashlykov, Soumya Batra, Prajjwal Bhargava, Shruti Bhosale, Dan Bikel, Lukas Blecher, Cristian Canton{-}Ferrer, Moya Chen, Guillem Cucurull, David Esiobu, Jude Fernandes, Jeremy Fu, Wenyin Fu, Brian Fuller, Cynthia Gao, Vedanuj Goswami, Naman Goyal, Anthony Hartshorn, Saghar Hosseini, Rui Hou, Hakan Inan, Marcin Kardas, Viktor Kerkez, Madian Khabsa, Isabel Kloumann, Artem Korenev, Punit~Singh Koura, Marie{-}Anne Lachaux, Thibaut Lavril, Jenya Lee, Diana Liskovich, Yinghai Lu, Yuning Mao, Xavier Martinet, Todor Mihaylov, Pushkar Mishra, Igor Molybog, Yixin Nie, Andrew Poulton, Jeremy Reizenstein, Rashi Rungta, Kalyan Saladi, Alan Schelten, Ruan Silva, Eric~Michael Smith, Ranjan Subramanian, Xiaoqing~Ellen Tan, Binh Tang, Ross Taylor, Adina Williams, Jian~Xiang Kuan, Puxin Xu, Zheng Yan, Iliyan Zarov, Yuchen Zhang, Angela Fan, Melanie Kambadur, Sharan Narang, Aur{\'{e}}lien Rodriguez, Robert Stojnic, Sergey Edunov,
  and Thomas Scialom. 2023.
\newblock Llama 2: Open foundation and fine-tuned chat models.
\newblock \emph{CoRR}, abs/2307.09288.

\bibitem[{Vaswani et~al.(2017)Vaswani, Shazeer, Parmar, Uszkoreit, Jones, Gomez, Kaiser, and Polosukhin}]{transformer}
Ashish Vaswani, Noam Shazeer, Niki Parmar, Jakob Uszkoreit, Llion Jones, Aidan~N. Gomez, Lukasz Kaiser, and Illia Polosukhin. 2017.
\newblock Attention is all you need.
\newblock In \emph{Advances in Neural Information Processing Systems 30: Annual Conference on Neural Information Processing Systems 2017, December 4-9, 2017, Long Beach, CA, {USA}}, pages 5998--6008.

\bibitem[{Vrandecic and Kr{\"{o}}tzsch(2014)}]{DBLP:journals/cacm/wikidata}
Denny Vrandecic and Markus Kr{\"{o}}tzsch. 2014.
\newblock Wikidata: a free collaborative knowledgebase.
\newblock \emph{Commun. {ACM}}, 57(10):78--85.

\bibitem[{Wen et~al.(2024)Wen, Wang, and Sun}]{mindmap}
Yilin Wen, Zifeng Wang, and Jimeng Sun. 2024.
\newblock Mindmap: Knowledge graph prompting sparks graph of thoughts in large language models.
\newblock In \emph{Proceedings of the 62nd Annual Meeting of the Association for Computational Linguistics (Volume 1: Long Papers), {ACL} 2024, Bangkok, Thailand, August 11-16, 2024}, pages 10370--10388. Association for Computational Linguistics.

\bibitem[{Yang et~al.(2015)Yang, Yih, He, Gao, and Deng}]{DBLP:journals/corr/DistMult}
Bishan Yang, Wen{-}tau Yih, Xiaodong He, Jianfeng Gao, and Li~Deng. 2015.
\newblock Embedding entities and relations for learning and inference in knowledge bases.
\newblock In \emph{{ICLR} (Poster)}.

\bibitem[{Yin et~al.(2023)Yin, Fu, Zhao, Li, Sun, Xu, and Chen}]{DBLP:journals/corr/mllm-survey}
Shukang Yin, Chaoyou Fu, Sirui Zhao, Ke~Li, Xing Sun, Tong Xu, and Enhong Chen. 2023.
\newblock A survey on multimodal large language models.
\newblock \emph{CoRR}, abs/2306.13549.

\bibitem[{Zhang et~al.(2022)Zhang, Chen, Yao, Chen, Zhu, Yu, Huang, Xu, Zhang, Xu, Yuan, Xiong, and Chen}]{DBLP:conf/sigir/NeuralKG}
Wen Zhang, Xiangnan Chen, Zhen Yao, Mingyang Chen, Yushan Zhu, Hongtao Yu, Yufeng Huang, Yajing Xu, Ningyu Zhang, Zezhong Xu, Zonggang Yuan, Feiyu Xiong, and Huajun Chen. 2022.
\newblock Neuralkg: An open source library for diverse representation learning of knowledge graphs.
\newblock In \emph{{SIGIR}}, pages 3323--3328. {ACM}.

\bibitem[{Zhang et~al.(2024{\natexlab{a}})Zhang, Chen, Fang, Lu, Li, Zhang, and Chen}]{KnowPAT}
Yichi Zhang, Zhuo Chen, Yin Fang, Yanxi Lu, Fangming Li, Wen Zhang, and Huajun Chen. 2024{\natexlab{a}}.
\newblock Knowledgeable preference alignment for llms in domain-specific question answering.
\newblock In \emph{Findings of the Association for Computational Linguistics, {ACL} 2024, Bangkok, Thailand and virtual meeting, August 11-16, 2024}, pages 891--904. Association for Computational Linguistics.

\bibitem[{Zhang et~al.(2024{\natexlab{b}})Zhang, Chen, Guo, Xu, Zhang, and Chen}]{KoPA-MM24}
Yichi Zhang, Zhuo Chen, Lingbing Guo, Yajing Xu, Wen Zhang, and Huajun Chen. 2024{\natexlab{b}}.
\newblock Making large language models perform better in knowledge graph completion.
\newblock In \emph{{ACM} Multimedia}, pages 233--242. {ACM}.

\bibitem[{Zhang et~al.(2023)Zhang, Li, Cui, Cai, Liu, Fu, Huang, Zhao, Zhang, Chen, Wang, Luu, Bi, Shi, and Shi}]{DBLP:journals/corr/Hallucination}
Yue Zhang, Yafu Li, Leyang Cui, Deng Cai, Lemao Liu, Tingchen Fu, Xinting Huang, Enbo Zhao, Yu~Zhang, Yulong Chen, Longyue Wang, Anh~Tuan Luu, Wei Bi, Freda Shi, and Shuming Shi. 2023.
\newblock Siren's song in the {AI} ocean: {A} survey on hallucination in large language models.
\newblock \emph{CoRR}, abs/2309.01219.

\bibitem[{Zhao et~al.(2023)Zhao, Zhou, Li, Tang, Wang, Hou, Min, Zhang, Zhang, Dong, Du, Yang, Chen, Chen, Jiang, Ren, Li, Tang, Liu, Liu, Nie, and Wen}]{DBLP:journals/corr/llmsurvey}
Wayne~Xin Zhao, Kun Zhou, Junyi Li, Tianyi Tang, Xiaolei Wang, Yupeng Hou, Yingqian Min, Beichen Zhang, Junjie Zhang, Zican Dong, Yifan Du, Chen Yang, Yushuo Chen, Zhipeng Chen, Jinhao Jiang, Ruiyang Ren, Yifan Li, Xinyu Tang, Zikang Liu, Peiyu Liu, Jian{-}Yun Nie, and Ji{-}Rong Wen. 2023.
\newblock A survey of large language models.
\newblock \emph{CoRR}, abs/2303.18223.

\end{thebibliography}

\appendix

\section{Details in {\benchmark} benchmark}
\subsection{Detailed Task Settings}
\label{sec:appendix:task_settings}
In this section, we provide a detailed description of the 9 task settings in the {\benchmark} benchmark. Note that we have 3 different granularity (EG/TG/SG) and 3 different levels (CLS/MC/DESC) in the {\benchmark} benchmark.
\begin{itemize}
    \item \textbf{Task 1: EG CLS.} This task needs the LLM to predict the true or false of a given question about whether a given embedding and an entity name are a pair.
    \item \textbf{Task 2: EG MC.} This task needs the LLM to select the true answer in the given 4 options to answer the question about a given entity embedding.
    \item \textbf{Task 3: EG DESC.} This task needs the LLM to generate a short entity name to answer what the entity is based on the input {\prompt}.
    \item \textbf{Task 4: TG CLS.} This task needs the LLM to predict the true or false of a given question about whether the {\prompt} of the triple is a positive one.
    \item \textbf{Task 5: TG MC.} This task needs the LLM to select the true answer in the given 4 options to complete the given query in the form of {\prompt}. The query can be a head prediction $(?, r, t)$, a relation prediction $(h, ?, t)$, or a tail prediction $(h, r, ?)$. Here, we denote $?$ as the missing entity/relation that needs to be completed by the model.
    \item \textbf{Task 6: TG DESC.} This task needs the LLM to generate the head/tail entity name and the relation to answer what the triple is based on the input {\prompt}.
    \item \textbf{Task 7: SG CLS.} This task needs the LLM to predict the true or false of a given question about whether the {\prompt} of the subgraph is a positive one.
    \item \textbf{Task 8: SG MC.} This task needs the LLM to select the true answer in the given 4 options to complete the given query in the form of {\prompt}. The query is a subgraph that removes a key entity, which should be predicted by the model.
    \item \textbf{Task 9: SG DESC.} This task needs the LLM to generate a paragraph to describe what the {\prompt} is in the given input. The subgraph is extracted from the KGs by random sampling.
\end{itemize}
Note that, for CLS tasks, an entity-short name pair/triple/subgraph sampled from the KG is regarded as a positive sample. We generate negative samples by randomly replacing the positive samples in a 1:1 manner. For the EG DESC and TG DESC tasks, the golden label for each entity and relation is their short name in the given KG. For the SG task, we employ GPT-3.5 to generate golden answers. The prompt template we used is presented in Figure~\ref{box:clean}. We manually verified the generated results and found that the generated golden answer is of acceptable quality and can be used to train models of around 7B.

\subsection{The Prompt Templates} 
\label{sec:appendix:prompt_template}

We present the prompt templates we used in the {\benchmark} benchmark in Figure~\ref{box:egcls} to Figure~\ref{box:sgdesc}. For each task, the instruction is consistent and the input would be changed by different data instances. We present one case for each task.

\begin{figure*}[h]
\centering
\begin{tcolorbox}[colback=green!10!white, colframe=green!60!black, title=Prompt for Golden Answer Generation]
Given several triples in an extracted subgraph from a knowledge graph, you need to organize them into text paragraphs to describe the information contained in this graph.
\\
The given triple:\\
<head\_1, relation\_1, tail\_1>\\
<head\_2, relation\_2, tail\_2>\\
…… \\
<head\_n, relation\_n, tail\_n>\\
Your Answer:
\end{tcolorbox}
\caption{The prompt template used to generate the golden answer of SG DESC task.}
\label{box:clean}
\end{figure*}

\begin{figure*}[t]
\centering
\begin{tcolorbox}[colback=green!10!white, colframe=green!60!black, title=GPT Evaluation Prompt Template]

Score the given model-generated text against the ground truth on a scale from 0 to 100, focusing on the alignment of meanings rather than the formatting.

\textbf{The ground truth text}: <Golden Label>

\textbf{The model output}: <Model Prediction>

Provide your score as a number and do not provide any other text in the response.
\end{tcolorbox}
\caption{The prompt template used for GPT evaluation.}
\label{box:scoring}
\end{figure*}

\section{Experiments}
\subsection{Experimental Details}
\label{sec:appendix:details}
In our experiments, we implement the training and evaluation process with PyTorch \cite{pytorch} and hugging-face transformers \cite{transformer} library. We train 3 epochs for each {\prompt} model with a fixed context length of 384. The batch size is set to 16. We tune the learning rate in $\{1e^{-4},3e^{-4},5e^{-4}\}$. 
\par For the structural encoders, we set the embedding dimensions of four different backbones to 512. We implement them using NeuralKG \cite{DBLP:conf/sigir/NeuralKG}, with a 3000 epoch training until coverage. The KG embedding methods (TransE/DistMult/RotatE/RGCN) are classic backbones to train structural embedding for a given KG. Besides, R-GCN \cite{DBLP:conf/esws/RGCN} employs a relational graph convolution layer for message aggregation in the KG. The training process is self-supervised. The training objective can be denoted as:
\begin{equation}
    \begin{aligned}
    \mathcal{L}=\frac{1}{|\mathcal{T}|}&\sum_{(h, r, t)\in \mathcal{T}}\Big(-\log\sigma(\gamma-\mathcal{F}(h, r, t))\\
    &-\sum_{i=1}^{K}p_i\log\sigma(\mathcal{F}(h_i',r_i',t_i')-\gamma)\Big)
    \end{aligned}
\end{equation}
where $(h, r, t)\in\mathcal{T}$ is a positive triple. $\sigma$ is the sigmoid function and $\gamma$ is a margin hyper-parameter. $p_i$ is the self-adversarial training weights proposed by RotatE \cite{DBLP:conf/iclr/RotatE}. $\mathcal{F}$ is the score function defined specifically by different methods. For example, the score function of TransE is:
\begin{equation}
    \mathcal{F}(h, r, t) = - ||\mathbf{h}+\mathbf{r}-\mathbf{t}||_1
\end{equation}
\par For the four kinds of adapters, we implement them with the following setting:
\begin{itemize}
    \item \textbf{FC}. It is implemented by a single linear layer in PyTorch in the form of $d_e\times d_{l}$, where $d_e$ is the structural embedding dimension and $d_l$ is the token embedding dimension of LLMs.
    \item \textbf{MLP}. It is implemented by several linear layers with ReLU \cite{relu} as an activation function. The intermediate dimension of the MLPs is $3\times d_e$. In most of the experiments, we use a two-layer MLP. In the scalability experiments, we explore deeper MLPs.
    \item \textbf{MoE}. We follow the implementation in XRec \cite{LLM4REC} for the MoE adapter layers. We set the expert number to 4 with adaptive gated fusion.
    \item \textbf{Qformer}. We follow the implementation in BLIP-2 \cite{BLIP2}. The number of transformer layers in Qformer is set to 2 with 2 attention heads. The readout layer is set to be a two-layer MLP.
\end{itemize}
Now we can explain why we chose these four kinds of adapters in our evaluation. Note that the paradigm of {\prompt} and MLLMs have certain ideas in common, which is \textbf{bridging heterogeneous information into LLM with adapters and employing texts as a core expression to solve different tasks}. Therefore, existing {\prompt} models are heavily informed by MLLMs.
\par Overall, they are all popular architectures used by existing methods. FC and MLP are fundamental neural networks used by GNP \cite{GNP-AAAI24} and KoPA \cite{KoPA-MM24}. MoE network is used by XRec, a work that attempts to inform LLMs with the structural information in user-item interaction graphs. Though different from the structural information in KGs, it is also worth a try. Qformer is a classic adapter widely used in Multi-modal LLMs such as BLIP-2, which is more complex than vanilla FC and MLP. Though no current work employs Qformer, we think Qformer is a representative design with amazing ideas. Therefore, we also evaluate it in our experiments.
\par The training process task about 20 to 30 minutes for EG and SG tasks in our experiment environment while TG takes about 4 hours. Some of our code implementation is under the help of AI assitant like ChatGPT.
\subsection{Evaluation Details}
\label{sec:appendix:evaluation}
In our evaluation protocol, we have several different metrics for different tasks. The CLS tasks and MC tasks have deterministic results, which can be measured by the quantitative accuracy metric. For the DESC, the situation becomes more complex. This is caused by the different settings in the DESC tasks. For EG DESC, we expect the model to generate a precise entity name. For TG DESC, we expect the model to generate the entities and relations properly in the given triple. For the SG DESC, the target is to create a paragraph to describe the given subgraph context. We can find that among these tasks, EG and TG require high accuracy and need to use Exact Match (EM) as an evaluation metric. SG, although also requires high accuracy, is not suitable for EM due to the generation of long text, so we adopt the current model of combining BLEU and GPTScore to evaluate the semantic similarity of the generated texts and the golden labels. The prompt template we used in the GPT evaluation is presented in Figure~\ref{box:scoring}.


\begin{table}[]
\centering
\caption{A further exploration on the influence of textual query in the TG MC task. We use the MLP as the adapter for {\prompt}.}
\label{table:notxt}
\resizebox{0.6\columnwidth}{!}{
\begin{tabular}{c|cc}
\toprule
 Method & w/ text & w/o text \\
\midrule
\textbf{TransE} & 86.51 & 49.09 \\
\textbf{DistMult} & 97.91 & 90.94 \\
\textbf{RotatE} & 83.74 & 83.51 \\
\textbf{R-GCN} & 94.59 & 78.45 \\
\bottomrule
\end{tabular}
}
\end{table}

\begin{table}[]
\caption{Results on LlaMA3.1-8B.}
\label{table:llama3.1}
\vspace{-8pt}
\centering
\resizebox{0.9\columnwidth}{!}{
\begin{tabular}{ccccc}
\toprule
\multicolumn{2}{c}{\textbf{Method}} & \textbf{Entity MC} & \textbf{Triple MC} & \textbf{Subgraph MC} \\
\midrule
\multirow{4}{*}{\textbf{FC}} & \textbf{TransE} & 51.86 & 88.26 & 49.35 \\
 & \textbf{DistMult} & 30.15 & 90.25 & 78.10 \\
 & \textbf{RotatE} & 35.67 & 72.65 & 46.28 \\
 & \textbf{R-GCN} & 40.04 & 89.12 & 65.61 \\
 \midrule
\multirow{4}{*}{\textbf{MLP}} & \textbf{TransE} & 94.83 & 93.04 & 85.05 \\
 & \textbf{DistMult} & 75.31 & 93.67 & 90.19 \\
 & \textbf{RotatE} & 45.65 & 88.84 & 91.59 \\
 & \textbf{R-GCN} & 67.81 & 93.78 & 90.93 \\
 \midrule
\multirow{4}{*}{\textbf{MoE}} & \textbf{TransE} & 44.16 & 89.94 & 78.56 \\
 & \textbf{DistMult} & 37.31 & 91.50 & 76.93 \\
 & \textbf{RotatE} & 61.00 & 91.60 & 81.83 \\
 & \textbf{R-GCN} & 46.24 & 90.54 & 68.84 \\
 \midrule
\multirow{4}{*}{\textbf{Q-former}} & \textbf{TransE} & 88.47 & 90.95 & 20.34 \\
 & \textbf{DistMult} & 75.21 & 93.14 & 44.18 \\
 & \textbf{RotatE} & 72.67 & 92.67 & 28.54 \\
 & \textbf{R-GCN} & 69.64 & 92.83 & 22.59 \\
 \bottomrule
\end{tabular}
}
\end{table}

\begin{table}[]
\caption{Results on LlaMA2-7B-chat.}
\label{table:llama2}
\vspace{-8pt}
\centering
\resizebox{0.9\columnwidth}{!}{
\begin{tabular}{cc|ccc}
\toprule
\multicolumn{2}{c|}{\textbf{Method}} & \textbf{Entity MC} & \textbf{Triple MC} & \textbf{Subgraph MC} \\
\midrule
\multirow{4}{*}{\textbf{FC}} & \textbf{TransE} & 34.72 & 46.77 & 43.29 \\
 & \textbf{DistMult} & 26.08 & 61.75 & 70.75 \\
 & \textbf{RotatE} & 23.00 & 67.21 & 52.66 \\
 & \textbf{R-GCN} & 18.18 & 64.62 & 50.91 \\
 \midrule
\multirow{4}{*}{\textbf{MLP}} & \textbf{TransE} & 83.38 & 91.32 & 69.38 \\
 & \textbf{DistMult} & 29.01 & 71.51 & 89.03 \\
 & \textbf{RotatE} & 72.37 & 72.88 & 55.73 \\
 & \textbf{R-GCN} & 37.06 & 89.01 & 79.19 \\
\midrule
\multirow{4}{*}{\textbf{MoE}} & \textbf{TransE} & 42.42 & 60.62 & 63.78 \\
 & \textbf{DistMult} & 42.37 & 61.05 & 71.64 \\
 & \textbf{RotatE} & 25.73 & 58.37 & 62.15 \\
 & \textbf{R-GCN} & 22.75 & 63.04 & 64.56 \\
\midrule
\multirow{4}{*}{\textbf{Q-former}} & \textbf{TransE} & 77.20 & 62.73 & 31.73 \\
 & \textbf{DistMult} & 37.04 & 89.93 & 40.80 \\
 & \textbf{RotatE} & 39.84 & 91.96 & 22.83 \\
 & \textbf{R-GCN} & 40.73 & 87.41 & 25.86 \\
\bottomrule
\end{tabular}
}
\end{table}

\subsection{The influence of textual query in TG-MC task.}
As we presented in the task definitions of {\benchmark}, we provide text-based descriptions of the given query in the TG MC task. For example, \textit{([MASK] | occupation | romanist)} in Figure~\ref{box:tgmc}. Besides, the options are also in the form of texts which means many questions can make text-based predictions without {\prompt} as well. To better investigate LLM's ability to understand {\prompt} on the task TG MC, we performed some additional implementations to validate the experiments in the absence of text. As shown in Figure~\ref{table:notxt}, it is obvious that the model performs better in the presence of text, because the query present in the form of text greatly simplifies LLM's understanding of the problem and makes the whole thing easier. But on the other hand, in the absence of text, the LLM still has some SKP comprehension and it can make the right choices relying on SKP alone.

\subsection{Additional results on LlaMA3.1 and LlaMA2}
\label{sec:appendix:llama}
We present more detailed experimental results in Figure~\ref{table:llama3.1} and Figure~\ref{table:llama2} about the {\benchmark} benchmark of LlaMA3.1-8B-Instruct and LlaMA2-7B-chat. These results are complementary to the universality experiment. We can observe that the model based on LlaMA3.1 performs relatively better in general compared to the model based on LlaMA2.

\begin{figure*}[htbp]
\centering
\begin{tcolorbox}[colback=green!10!white, colframe=green!60!black, title=Entity Classification (EG-CLS)]
<Structural Knowledge Prompting $\mathcal{S}_{e}$ >\\
\textbf{\#\#\# Instruction}: Given an entity embedding in the front and answer the following question. \\
\textbf{\#\#\# Question:} Is this entity Bonnie Owens?\\
\textbf{\#\#\# Output:} Yes.
\end{tcolorbox}
\caption{The prompt template for EG-CLS.}
\label{box:egcls}
\end{figure*}

\begin{figure*}
\begin{tcolorbox}[colback=green!10!white, colframe=green!60!black, title=Entity Multiple Choice (EG-MC)]

<Structural Knowledge Prompting $\mathcal{S}_{e}$ >\\ \textbf{\#\#\# Instruction}: Given an entity embedding in the front and select the true answer.\\
\textbf{\#\#\# Question:} Which is the possible type of entity? A. bell tower B. congestive heart failure C. academic major D. wall hanging\\
\textbf{\#\#\# Output:} Your Answer is: D
\end{tcolorbox}
\caption{The prompt template for EG-MC.}
\label{box:egmc}
\end{figure*}

\begin{figure*}
\begin{tcolorbox}[colback=green!10!white, colframe=green!60!black, title=Entity Description (EG-DESC)]

<Structural Knowledge Prompting $\mathcal{S}_{e}$ >\\ \textbf{\#\#\# Instruction}: Given an entity embedding in the front and describe this entity.\\
\textbf{\#\#\# Output:} Bonnie Owens. \#\#\# Dessctiption: Bonnie Owens , born Bonnie Campbell, was an American country music singer who was married to Buck Owens and later Merle Haggard.
\end{tcolorbox}
\caption{The prompt template for EG-DESC.}
\label{box:egdesc}
\end{figure*}

\begin{figure*}
\begin{tcolorbox}[colback=green!10!white, colframe=green!60!black, title=Triple Classification (TG-CLS)]

<Structural Knowledge Prompting $\mathcal{S}_{h}$, $\mathcal{S}_{r}$, $\mathcal{S}_{t}$ >\\ \textbf{\#\#\# Instruction}: Given the embeddings of an knowledge triple in the front. Please answer the following question.\\
\textbf{\#\#\# Question: } Is this a positive triple? \\
\textbf{\#\#\# Output:} No
\end{tcolorbox}
\caption{The prompt template for TG-CLS.}
\label{box:tgcls}
\end{figure*}

\begin{figure*}
\begin{tcolorbox}[colback=green!10!white, colframe=green!60!black, title=Triple Multiple Choice (TG-MC)]

<Structural Knowledge Prompting $\mathcal{S}_{r}$, $\mathcal{S}_{t}$ > \# Note that the masked answer does not appear in {\prompt}
\\ \textbf{\#\#\# Instruction}: Given the embeddings of a query and four candidates in the front. Select a correct answer to fill the [MASK] and complete the triple.\\
\textbf{\#\#\# Question:} ([MASK] | occupation | romanist) A. László András B. Rebekah Brooks C. Franz Konwitschny D. Francisco Rodríguez Marín  \\
\textbf{\#\#\# Output:} A
\end{tcolorbox}
\caption{The prompt template for TG-MC.}
\label{box:tgmc}
\end{figure*}

\begin{figure*}
\begin{tcolorbox}[colback=green!10!white, colframe=green!60!black, title=Triple Description (TG-DESC)]

<Structural Knowledge Prompting $\mathcal{S}_{h}$, $\mathcal{S}_{r}$, $\mathcal{S}_{t}$ >\\ \textbf{\#\#\# Instruction}: Given the embeddings of an knowledge triple in the front and describe the head entity, relation, and tail entity of the triple.\\
\textbf{\#\#\# Output:} Billy Idol\#\#\#languages spoken, written, or signed\#\#\#English
\end{tcolorbox}
\caption{The prompt template for TG-DESC.}
\label{box:tgdesc}
\end{figure*}

\begin{figure*}
\begin{tcolorbox}[colback=green!10!white, colframe=green!60!black, title=Subgraph Classification (SG-CLS)]

<Structural Knowledge Prompting $\mathcal{S}_{h},\mathcal{S}_{r_1},\mathcal{S}_{t_1}, \cdots, \mathcal{S}_{r_k},\mathcal{S}_{t_k}$>\\ \textbf{\#\#\# Instruction}: Given embeddings from a subgraph in the front and answer the following question.\\
\textbf{\#\#\# Question:} Is there any anomaly in this subgraph?\\
\textbf{\#\#\# Output:} No
\end{tcolorbox}
\caption{The prompt template for SG-CLS.}
\label{box:sgcls}
\end{figure*}

\begin{figure*}
\begin{tcolorbox}[colback=green!10!white, colframe=green!60!black, title=Subgraph Multiple Choice (SG-MC)]

<Structural Knowledge Prompting $\mathcal{S}_{r_1},\mathcal{S}_{t_1}, \cdots, \mathcal{S}_{r_k},\mathcal{S}_{t_k}$> \# The center entity does not appear.\\ \textbf{\#\#\# Instruction}: Given an entity embedding in the front and select the true answer.\\
\textbf{\#\#\# Question:} Which is the center entity descripted by this subgraph? A. The Lord of the Rings: The Fellowship of the Ring B. Christian Reimers C. Harry Fett D. Manuel Acevedo 
\\
\textbf{\#\#\# Output:} Your Answer is: D
\end{tcolorbox}
\caption{The prompt template for SG-MC.}
\label{box:sgmc}
\end{figure*}

\begin{figure*}
\begin{tcolorbox}[colback=green!10!white, colframe=green!60!black, title=Subgraph Description (SG-DESC)]

<Structural Knowledge Prompting $\mathcal{S}_{h},\mathcal{S}_{r_1},\mathcal{S}_{t_1}, \cdots, \mathcal{S}_{r_k},\mathcal{S}_{t_k}$>\\ \textbf{\#\#\# Instruction}: Given embeddings from a subgraph in the front and answer the following question.\\
\textbf{\#\#\# Output:} Dominic Howard is a musician from the United Kingdom, known for playing the drum kit in the genre of alternative rock. He primarily works in the field of music.
\end{tcolorbox}
\caption{The prompt template for SG-DESC.}
\label{box:sgdesc}
\end{figure*}

\end{document}